\documentclass[11pt]{article}
\usepackage[a4paper]{geometry}
\usepackage{clic2017}
\usepackage{times}
\usepackage{url}
\usepackage{latexsym}
\usepackage{multirow}
\usepackage{multicol}
\usepackage{graphicx}
\usepackage{lingmacros} 

\title{Italian Event Detection Goes Deep Learning}

\author{Tommaso Caselli \\
  CLCG, Rijksuniversiteit Groningen 
  Oude Kijk in't Jaatsraaat, 26 \\
  9712 EK Groningen (NL) \\
  {\tt t.caselli@\{rug.nl\}\{gmail.com\}}}

\date{}

\date{}

\begin{document}
\maketitle
\begin{abstract}
  \textbf{English.} This paper reports on a set of experiments with different word embeddings to initialize a state-of-the-art Bi-LSTM-CRF network for event detection and classification in Italian, following the EVENTI evaluation exercise. The network obtains a new state-of-the-art result by improving the F1 score for detection of 1.3 points, and of 6.5 points for classification, by using a single step approach. The results also provide further evidence that embeddings have a major impact on the performance of such architectures.
\end{abstract}

\begin{abstract-alt}
 \textrm{\bf{Italiano.}} Questo contributo descrive una serie di esperimenti con diverse rappresentazioni distribuzionali di parole (\textit{word embeddings}) per inizializzare una rete neurale stato dell'arte di tipo Bi-LSTM-CRF per il riconoscimento e la classificazione di eventi in italiano, in base all'esercizio di valutazione EVENTI. La rete migliora lo stato dell'arte di 1.3 punti di F1 per il riconoscimento, e di 6.5 punti per la classificazione, affrontando il compito in un unico sistema. L'analisi dei risultati fornisce ulteriore supporto al fatto che le rappresentazioni distribuzionali di parole hanno un impatto molto alto nei risultati di queste architetture.
\end{abstract-alt}

\section{Introduction}

Current societies are exposed to a continuous flow of information that results in a large production of data (e.g. news articles, micro-blogs, social media posts, among others), at different moments in time. In addition to this, the consumption of information has dramatically changed: more and more people directly access information through social media platforms (e.g. Facebook and Twitter), and are less and less exposed to a diversity of perspectives and opinions. The combination of these factors may easily result in \textit{information overload} and impenetrable ``\textit{filter bubbles}''. Events, i.e. things that happen or hold as true in the world, are the basic components of such data stream. Being able to correctly identify and classify them plays a major role to develop robust solutions to deal with the current stream of data (e.g. the storyline framework~\cite{vossen2015storylines}), as well to improve the performance of many Natural Language Processing (NLP) applications such as automatic summarization and question answering (Q.A.). 

Event detection and classification has seen a growing interest in the NLP community thanks to the availability of annotated corpora~\cite{ACE-event-2005,pustetal03-IWCS,W16-5706,CYBULSKA+VOSSEN:2014} and evaluation campaigns~\cite{verhagen-EtAl:2007:SemEval-2007,verhagen2010semeval,uzzanametal2013,bethard2015semeval,bethard2016semeval,minard2015semeval}. In the context of the 2014 EVALITA Workshop, the EVENTI evaluation exercise~\cite{casellietal-evalita2014}\footnote{\url{https://sites.google.com/site/eventievalita2014/}} was organized to promote research in Italian Temporal Processing, of which event detection and classification is a core subtask. 

Since the EVENTI campaign, there has been a lack of further research, especially in the application of deep learning models to this task in Italian. The contributions of this paper are the followings: i.) the adaptation of a state-of-the-art sequence to sequence (seq2seq) neural system to event detection and classification for Italian in a single step approach; ii.) an investigation on the quality of existing Italian word embeddings for this task; iii.) a comparison against a state-of-the-art discrete classifier. The pre-trained models and scripts running the system (or re-train it) are publicly available.~\footnote{\url{https://github.com/tommasoc80/Event_detection_CLiC-it2018}}.

\begin{table*}[!t]
\begin{minipage}[!th]{0.45\textwidth}
\centering
\small
\begin{tabular}{lrrr}
POS & Training & Dev. & Test \\ \hline
Noun &  6,710 & 111 & 1,499 \\
Verb & 11,269 & 193 & 2,426 \\
Adjective & 610 & 9 & 118  \\
Preposition & 146 & 1 & 25\\
\hline
Overall Event Tokens & 18,735 & 314 & 4,068  \\
\end{tabular}
 \caption{\label{tab:pos} Distribution of the event mentions per POS per token in all datasets of the EVENTI corpus.}
\end{minipage}
\hfill
\begin{minipage}[!th]{0.45\textwidth}
\centering
\small
\begin{tabular}{lrrr}
Class & Training & Dev. & Test  \\ \hline
OCCURRENCE & 9,041 & 162 & 1,949\\
ASPECTUAL & 446 & 14 & 107 \\
I\_STATE & 1,599 & 29 & 355 \\
I\_ACTION & 1,476 & 25 & 357 \\
PERCEPTION & 162 & 2 & 37 \\
REPORTING & 714 & 8 & 149 \\
STATE  & 4,090 & 61 & 843 \\
\hline
Overall Events & 17,528 & 301 & 3,798  \\
\end{tabular}
 \caption{\label{tab:class} Distribution of the event mentions per class in all datasets of the EVENTI corpus.}
\end{minipage}
\end{table*}

\section{Task Description}

We follow the formulation of the task as specified in the EVENTI exercise: determine the extent and the class of event mentions in a text, according to the It-TimeML $<$EVENT$>$ tag definition (Subtask B in EVENTI). 

In EVENTI, the tag $<$EVENT$>$ is applied to every linguistic expression denoting a situation that happens or occurs, or a state in which something obtains or holds true, regardless of the specific parts-of-speech that may realize it. EVENTI distinguishes between single token and multi-tokens events, where the latter are restricted to specific cases of eventive multi-word expressions in lexicographic dictionaries (e.g. ``\textit{fare le valigie}'' [to pack]), verbal periphrases (e.g. ``\textit{(essere) in grado di}'' [(to be) able to]; ``\textit{c'\`e}'' [there is]), and named events (e.g. ``\textit{la strage di Beslan}'' [Beslan school siege]).

Each event is further assigned to one of 7 possible classes, namely: OCCURRENCE, ASPECTUAL, PERCEPTION, REPORTING, I(NTESIONAL)\_STATE, I(NTENSIONAL)\_ACTION, and STATE. These classes are derived from the English TimeML Annotation Guidelines~\cite{pustejovsky2003timeml}. The TimeML event classes distinguishes with respect to other classifications, such as ACE~\cite{ACE-event-2005} or FrameNet~\cite{framenet}, because they expresses relationships the target event participates in (such as factual, evidential, reported, intensional) rather than semantic categories denoting the meaning of the event. This means that the EVENT classes are assigned by taking into account both the semantic and the syntactic context of occurrence of the target event. Readers are referred to the EVENTI Annotation Guidelines for more details\footnote{\url{https://sites.google.com/site/eventievalita2014/file-cabinet}}.

\subsection{Dataset}

The EVENTI corpus consists of three datasets: the Main Task training data, the Main task test data, and the Pilot task test data. The Main Task data are on contemporary news articles, while the Pilot Task on historical news articles. For our experiments, we focused only on the Main Task. In addition to the training and test data, we have created also a Main Task development set by excluding from the training data all the articles that composed the test data of the Italian dataset at the SemEval 2010 TempEval-2 campaign~\cite{verhagen2010semeval}. The new partition of the corpus results in the following distribution of the $<$EVENT$>$ tag: i) 17,528 events in the training data, of which 1,207 are multi-token mentions; ii.) 301 events in the development set, of which 13 are multi-token mentions; and finally, iii.) 3,798 events in the Main task test, of which 271 are multi-token mentions. 

Tables~\ref{tab:pos} and ~\ref{tab:class} report, respectively, the distribution of the events per token part-of speech (POS) and per event class. Not surprisingly, verbs are the largest annotated category, followed by nouns, adjectives, and prepositional phrases. Such a distribution reflects both a kind of ``natural'' distribution of the realization of events in an Indo-european language, and, at the same time, specific annotation choices. For instance, adjectives have been annotated only when in a predicative position and when introduced by a copula or a copular construction. As for the classes, OCCURRENCE and STATE represent the large majority of all events, followed by the intensional ones (I\_STATE and I\_ACTION), expressing some factual relationship between the target events and their arguments, and finally the others (REPORTING, ASPECTUAL, and PERCEPTION).

\section{System and Experiments}

We adapted a publicly available Bi-LSTM network with a CRF classifier as last layer~\cite{reimers-gurevych:2017:EMNLP2017}.~\footnote{\url{https://github.com/UKPLab/emnlp2017-bilstm-cnn-crf}} \cite{reimers-gurevych:2017:EMNLP2017} demonstrated that word embeddings, among other hyper-parameters, have a major impact on the performance of the network, regardless of the specific task.   
On the basis of these experimental observations, we decided to investigate the impact of different Italian word embeddings for the Subtask B Main Task of the EVENTI exercise. We thus selected 5 word embeddings for Italian to initialize the network, differentiating one with respect to each other either for the representation model used (\texttt{word2vec} \textit{vs}. GloVe; CBOW \textit{vs}. skip-gram), dimensionality (300 \textit{vs}. 100), or corpora used for their generation (Italian Wikipedia \textit{vs}. crawled web document \textit{vs}. large textual corpora or archives):

\begin{table*}[!t]
\centering
\small
\begin{tabular}{lrrrrrrrr}

 & \multicolumn{4}{r}{Strict Evaluation} & \multicolumn{4}{r}{Relaxed Evaluation} \\ \hline
Embedding Parameter & R & P & F1 & F1-class & R & P & F1 & F1-class  \\ \hline
Berardi2015\_w2v & 0.868 & 0.868 & 0.868 & 0.705 & 0.892 & 0.892 & 0.892 & 0.725 \\
Berardi2015\_Glove & 0.848 & 0.872 & 0.860 & 0.697 & 0.870 & 0.895 & 0.882 & 0.714 \\
Fastext-It &  \bf 0.897	& 0.863 & \bf 0.880 & \bf 0.736 & \bf 0.921	& 0.887 & \bf 0.903 & \bf 0.756 \\
ILC-ItWack & 0.831 &  \bf 0.884 & 0.856 & 0.702 &  0.860 & \bf 0.914 & 0.886 & 0.725 \\
DH-FBK\_100 & 0.855 & 0.859 & 0.857 & 0.685 & 0.881 & 0.885 & 0.883 & 0.705 \\
\hline
\hline 
FBK-HLT@EVENTI 2014 & 0.850 & \textit{0.884} & 0.867 & 0.671 & 0.868 & 0.902 & 0.884 & 0.685  \\ 

\end{tabular}
 \caption{\label{tab:subtask-all} Results for Bubtask B Main Task - Event detection and classification.}
\end{table*}

\begin{itemize}
\item Berardi2015\_w2v~\cite{berardi2015word}: 300 dimension word embeddings generated using the \texttt{word2vec}~\cite{mikolov2013distributed} skip-gram model~\footnote{Parameters: negative sampling 10, context window 10} 
from the Italian Wikipedia;
\item Berardi2015\_glove~\cite{berardi2015word}: 300 dimensions word embeddings generated using the GloVe model~\cite{pennington2014glove} from the Italian Wikipedia\footnote{Berardi2015\_w2v and Berardi2015\_glove uses a 2015 dump of the Italian Wikipedia};
\item Fastext-It: 300 dimension word embeddings from the Italian Wikipedia~\footnote{Wikipedia dump not specified.} obtained using Bojanovsky's skip-gram model representation~\cite{bojanowski2016enriching}, where each word is represented as a bag of character n-grams~\footnote{\url{https://github.com/facebookresearch/fastText/blob/master/pretrained-vectors.md}}; 
\item ILC-ItWack~\cite{cimino2016building}: 300 dimension word embeddings generated by using the \texttt{word2vec} CBOW model~\footnote{Parameters: context window 5.} 
from the ItWack corpus;
\item DH-FBK\_100~\cite{tonelliimpact}: 100 dimension word and phrase embeddings, generated using the \texttt{word2vec} and \texttt{phrase2vec} models, from 1.3 billion word corpus (Italian Wikipedia, OpenSubtitles2016~\cite{lison2016opensubtitles2016}, PAISA corpus~\footnote{\url{http://www.corpusitaliano.it/}}, and the Gazzetta Ufficiale). 
\end{itemize}

As for the other parameters, the network maintains the optimized configurations used for the event detection task for English~\cite{reimers-gurevych:2017:EMNLP2017}: two LSTM layers of 100 units each, \textit{Nadam} optimizer, variational dropout (0.5, 0.5), with gradient normalization ($\tau$ = 1), and batch size of 8. Character-level embeddings, learned using a Convolutional Neural Network (CNN)~\cite{P16-1101}, are concatenated with the word embedding vector to feed into the LSTM network. Final layer of the network is a CRF classifier.

Evaluation is conducted using the EVENTI evaluation framework. Standard Precision, Recall, and F1 apply for the event detection. Given that the extent of an event tag may be composed by more than one tokens, systems are evaluated both for strict match, i.e. one point only if all tokens which compose an $<$EVENT$>$ tag are correctly identified, and relaxed match, i.e. one point for any correct overlap between the system output and the reference gold data. The classification aspect is evaluated using the F1-attribute score~\cite{uzzanametal2013}, that captures how well a system identify both the entity (extent) and attribute (i.e. class) together.

We approached the task in a single-step by detecting and classifying event mentions at once rather than in the standard two step approach, i.e. detection first and classification on top of the detected elements. The task is formulated as a seq2seq problem, by converting the original annotation format into an BIO scheme (Beginning, Inside, Outside), with the resulting alphabet being B-\textit{class\_label}, I-\textit{class\_label} and O. Example~\ref{example-rec-class-event} below illustrates a simplified version of the problem for a short sentence:

\enumsentence{
\begin{small}
\begin{tabbing}
 \ \= \underline{input \ \ } \  \= \underline{\ problem\ \ \ \ \ \ \ \ \ \ \ \ \ \ \ \ \ \ \ \ \ \ \ \ \ \ \ \ \ \ \ \ \ \ } \ \ \ \ \ \ \= \underline{solution}\\
\> Marco  \> (B-STATE $|$ I-STATE $|$ \dots $|$ O)  \>  O\\
\> pensa  \>  (B-STATE $|$ I-STATE $|$ \dots $|$ O)  \> B-ISTATE\\
\> di  \>  (B-STATE $|$ I-STATE $|$ \dots $|$ O)  \> O\\
\> andare \>  (B-STATE $|$ I-STATE $|$ \dots $|$ O)  \> B-OCCUR\\
\> a     \> (B-STATE $|$ I-STATE $|$ \dots $|$ O)  \> O\\
\> casa \>  (B-STATE $|$ I-STATE $|$ \dots $|$ O)  \> O\\
\> .     \>  (B-STATE $|$ I-STATE $|$ \dots $|$ O)  \> O\\
\end{tabbing}
\end{small}}\label{example-rec-class-event}

\subsection{Results and Discussion}

\begin{figure}
\begin{center}
\includegraphics[width=0.5\textwidth]{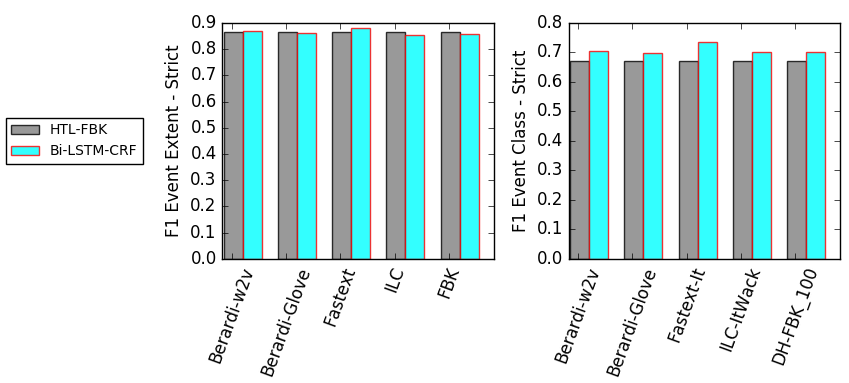}
\caption{Plots of F1 scores of the Bi-LSTM-CRF systems against the FBK-HLT system for Event Extent (left side) and Event Class (right side). F1 scores refers to the }
\label{fig:volumes}
\end{center}
\end{figure}

Results for the experiments are illustrated in Table~\ref{tab:subtask-all}. We also report the results of the best system that participated at EVENTI Subtask B, FBK-HLT~\cite{mirza2014fbk}. FBK-HLT is a cascade of two SVM classifiers (one for detection and one for classification) based on rich linguistic features. Figure~\ref{fig:volumes} plots charts comparing F1 scores of the network initialized with each of the five embeddings against the FBK-HLT system for the event detection and classification tasks, respectively.

The results of the Bi-LSTM-CRF network are varied in both evaluation configurations. The differences are mainly due to the embeddings used to initialize the network. The best embedding configuration is Fastext-It that differentiate from all the others for the approach used for generating the embeddings. Embedding's  dimensionality impacts on the performances supporting the findings in ~\cite{reimers-gurevych:2017:EMNLP2017}, but it seems that the quantity (and variety) of data used to generate the embeddings can have a mitigating effect, as shown by the results of the DH-FBK-100 configuration (especially in the classification subtask, and in the Recall scores for the event extent subtask). Coverage of the embeddings (and consequenlty, tokenization of the dataset and the embeddings) is a further aspect to keep into account, but it seems to have a minor impact with respect to dimensionality. It turns out that~\cite{berardi2015word}'s embeddings are those suffering the most from out of vocabulary (OVV) tokens (2.14\% and 1.06\% in training, 2.77\% and 1.84\% in test for the \texttt{word2vec} model and GloVe, respectively) with respect to the others. However, they still outperform DH-FBK\_100 and ILC-ItWack, whose OVV are much lower (0.73\% in training and 1.12\% in test for DH-FBK\_100; 0.74\% in training and 0.83\% in test for ILC-ItWack).  

The network obtains the best F1 score, both for detection (F1 of 0.880 for strict evaluation and 0.903 for relaxed evaluation with Fastext-It embeddings) and for classification (F1-class of 0.756 for strict evaluation, and 0.751 for relaxed evaluation with Fastext-It embeddings). Although FBK-HLT suffers in the classification subtask, it qualifies as a highly competitive system for the detection subtask. By observing the strict F1 scores, FBK-HLT beats three  configurations (DH-FBK-100, ILC-ItWack, Berardi2015\_Glove)~\footnote{$p$-value $<$ 0.005 only against Berardi2015\_Glove and DH-FBK-100, with McNemar's test.}, almost equals one (Berardi2015\_w2v)~\footnote{$p$-value $>$ 0.005 with McNemar's test.}, and it is outperformed only by one (Fastext-It)~\footnote{$p$-value $<$ 0.005 with McNemar's test.}. In the relaxed evaluation setting, DH-FBK-100 is the only configuration that does not beat FBK-HLT (although the difference is only 0.001 point). Nevertheless, it is remarkable to observe that FBK-HLT has a very high Precision (0.902, relaxed evaluation mode), that is overcome by only one embedding configuration, ILC-ItWack. The results also indicates that word embeddings have a major contribution on Recall, supporting observations that distributed representations have better generalization capabilities than discrete feature vectors. This is further supported by the fact that these results are obtained using a single step approach, where the network has to deal with a total of 15 possible different labels.

We further compared the outputs of the best model, i.e. Fastext-It, against FBK-HLT. As for the event detection subtask, we have adopted an event-based analysis rather than a token based one, as this will provide better insights on errors concerning multi-token events and event parts-of-speech (see Table~\ref{tab:pos} for reference).~\footnote{Note that POS are manually tagged for events, not for their components.} By analyzing the True Positives, we observe that the Fastext-It model has better performances than FBK-HLT with nouns (77.78\% \textit{vs}. 65.64\%, respectively) and prepositional phrases (28.00\% \textit{vs}. 16.00\%, respectively). Performances are very close for verbs (88.04\% \textit{vs}. 88.49\%, respectively) and adjectives (80.50\% \textit{vs}. 79.66\%, respectively). These results, especially those for prepositional phrases, indicates that the Bi-LSTM-CRF network structure and embeddings are also much more robust at detecting multi-tokens instances of events, and difficult realizations of events, such as nouns.

Concerning the classification, we focused on the mismatches between correctly identified events (extent layer) and class assignment. The Fastext-It model wrongly assigns the class to only 557 event tokens compared to the 729 cases for FBK-HLT. The distribution of the class errors, in terms of absolute numbers, is the same between the two systems, with the top three wrong classes being, in both cases, OCCURRENCE, I\_ACTION and STATE. OCCURRENCE, not surprisingly, is the class that tends to be assigned more often by both systems, being also the most frequent. However, if FBK-HLT largely overgeneralizes OCCURRENCE (59.53\% of all class errors), this corresponds to only one third of the errors (37.70\%) in the Bi-LSTM-CRF network. Other notable differences concern I\_ACTION (27.82\% of errors for the Bi-LSTM-CRF \textit{vs}. 17.28\% for FBK-HLT), STATE (8.79\% for the Bi-LSTM-CRF \textit{vs}. 15.22\% for FBK-HLT) and REPORTING (7.89\% for the Bi-LSTM-CRF \textit{vs}. 2.33\% for FBK-HLT) classes.

\section{Conclusion and Future Work}

This paper has investigated the application of different word embeddings for the initialization of a state-of-the-art Bi-LSTM-CRF network to solve the event detection and classification task in Italian, according to the EVENTI exercise. We obtained new state-of-the-art results using the Fastext-It embeddings, and improved the F1-class score of 6.5 points in strict evaluation mode. As for the event detection subtask, we observe a limited improvement (+1.3 points in strict F1), mainly due to gains in Recall. Such results are extremely positive as the task has been modeled in a single step approach, i.e. detection and classification at once, for the first time in Italian. Further support that embeddings have a major impact in the performance of neural architectures is provided, as the variations in performance of the Bi-LSMT-CRF models show. This is due to a combination of factors such as dimensionality, (raw) data, and the method used for generating the embeddings. 

Future work should focus on the development of embeddings that move away from the basic word level, integrating extra layers of linguistic analysis (e.g. syntactic dependencies)~\cite{komninos2016dependency}, that have proven to be very powerful for the same task in English. 


\section*{Acknowledgments}
The author wants to thank all researchers and research groups who made available their word embeddings and their code. Sharing is caring.

\bibliographystyle{acl}
\bibliography{eventibiblio}

\newcommand{\noop}[1]{}
\begin{thebibliography}{}

\bibitem[\protect\citename{Baker \bgroup et al.\egroup }1998]{framenet}
Collin~F Baker, Charles~J Fillmore, and John~B Lowe.
\newblock 1998.
\newblock The berkeley framenet project.
\newblock In {\em Proceedings of the 17th international conference on
  Computational linguistics-Volume 1}, pages 86--90. Association for
  Computational Linguistics.

\bibitem[\protect\citename{Berardi \bgroup et al.\egroup
  }2015]{berardi2015word}
Giacomo Berardi, Andrea Esuli, and Diego Marcheggiani.
\newblock 2015.
\newblock Word embeddings go to italy: A comparison of models and training
  datasets.
\newblock In {\em IIR}.

\bibitem[\protect\citename{Bethard \bgroup et al.\egroup
  }2015]{bethard2015semeval}
Steven Bethard, Leon Derczynski, Guergana Savova, James Pustejovsky, and Marc
  Verhagen.
\newblock 2015.
\newblock Semeval-2015 task 6: Clinical tempeval.
\newblock In {\em Proceedings of the 9th International Workshop on Semantic
  Evaluation (SemEval 2015)}, pages 806--814.

\bibitem[\protect\citename{Bethard \bgroup et al.\egroup
  }2016]{bethard2016semeval}
Steven Bethard, Guergana Savova, Wei-Te Chen, Leon Derczynski, James
  Pustejovsky, and Marc Verhagen.
\newblock 2016.
\newblock Semeval-2016 task 12: Clinical tempeval.
\newblock In {\em Proceedings of the 10th International Workshop on Semantic
  Evaluation (SemEval-2016)}, pages 1052--1062.

\bibitem[\protect\citename{Bojanowski \bgroup et al.\egroup
  }2016]{bojanowski2016enriching}
Piotr Bojanowski, Edouard Grave, Armand Joulin, and Tomas Mikolov.
\newblock 2016.
\newblock Enriching word vectors with subword information.
\newblock {\em arXiv preprint arXiv:1607.04606}.

\bibitem[\protect\citename{Caselli \bgroup et al.\egroup
  }2014]{casellietal-evalita2014}
T.~Caselli, R.~Sprugnoli, M.~Speranza, and M.~Monachini.
\newblock 2014.
\newblock Eventi.{E}{V}aluation of {E}vents and {T}emporal {I}{N}formation at
  {E}valita 2014.
\newblock In C.~Bosco, F.~Dell’Orletta, S.~Montemagni, and M.~Simi, editors,
  {\em Evaluation of Natural Language and Speech Tools for Italian}, volume~1,
  pages 27–--34. Pisa University Press.

\bibitem[\protect\citename{Cimino and Dell'Orletta}2016]{cimino2016building}
Andrea Cimino and Felice Dell'Orletta.
\newblock 2016.
\newblock Building the state-of-the-art in pos tagging of italian tweets.
\newblock In {\em CLiC-it/EVALITA}.

\bibitem[\protect\citename{Cybulska and Vossen}2014]{CYBULSKA+VOSSEN:2014}
Agata Cybulska and Piek Vossen.
\newblock 2014.
\newblock Using a sledgehammer to crack a nut? {L}exical diversity and event
  coreference resolution.
\newblock In {\em Proceedings of the 9th Language Resources and Evaluation
  Conference (LREC2014)}, Reykjavik, Iceland, May 26-31.

\bibitem[\protect\citename{Komninos and Manandhar}2016]{komninos2016dependency}
Alexandros Komninos and Suresh Manandhar.
\newblock 2016.
\newblock Dependency based embeddings for sentence classification tasks.
\newblock In {\em Proceedings of the 2016 Conference of the North American
  Chapter of the Association for Computational Linguistics: Human Language
  Technologies}, pages 1490--1500.

\bibitem[\protect\citename{LDC}2005]{ACE-event-2005}
LDC.
\newblock 2005.
\newblock Ace (automatic content extraction) english annotation guidelines for
  events ver. 5.4.3 2005.07.01.
\newblock In {\em Linguistic Data Consortium}.

\bibitem[\protect\citename{Lison and
  Tiedemann}2016]{lison2016opensubtitles2016}
Pierre Lison and J{\"o}rg Tiedemann.
\newblock 2016.
\newblock Opensubtitles2016: Extracting large parallel corpora from movie and
  tv subtitles.

\bibitem[\protect\citename{Ma and Hovy}2016]{P16-1101}
Xuezhe Ma and Eduard Hovy.
\newblock 2016.
\newblock End-to-end sequence labeling via bi-directional lstm-cnns-crf.
\newblock In {\em Proceedings of the 54th Annual Meeting of the Association for
  Computational Linguistics (Volume 1: Long Papers)}, pages 1064--1074.
  Association for Computational Linguistics.

\bibitem[\protect\citename{Mikolov \bgroup et al.\egroup
  }2013]{mikolov2013distributed}
Tomas Mikolov, Ilya Sutskever, Kai Chen, Greg~S Corrado, and Jeff Dean.
\newblock 2013.
\newblock Distributed representations of words and phrases and their
  compositionality.
\newblock In {\em Advances in neural information processing systems}, pages
  3111--3119.

\bibitem[\protect\citename{Minard \bgroup et al.\egroup
  }2015]{minard2015semeval}
Anne-Lyse Minard, Manuela Speranza, Eneko Agirre, Itziar Aldabe, Marieke van
  Erp, Bernardo Magnini, German Rigau, Ruben Urizar, and Fondazione~Bruno
  Kessler.
\newblock 2015.
\newblock Semeval-2015 task 4: Timeline: Cross-document event ordering.
\newblock In {\em Proceedings of the 9th International Workshop on Semantic
  Evaluation (SemEval 2015)}, pages 778--786.

\bibitem[\protect\citename{Mirza and Minard}2014]{mirza2014fbk}
Paramita Mirza and Anne-Lyse Minard.
\newblock 2014.
\newblock Fbk-hlt-time: a complete italian temporal processing system for
  eventi-evalita 2014.
\newblock In {\em Fourth International Workshop EVALITA 2014}, pages 44--49.

\bibitem[\protect\citename{O'Gorman \bgroup et al.\egroup }2016]{W16-5706}
Tim O'Gorman, Kristin Wright-Bettner, and Martha Palmer.
\newblock 2016.
\newblock Richer event description: Integrating event coreference with
  temporal, causal and bridging annotation.
\newblock In {\em Proceedings of the 2nd Workshop on Computing News Storylines
  (CNS 2016)}, pages 47--56. Association for Computational Linguistics.

\bibitem[\protect\citename{Pennington \bgroup et al.\egroup
  }2014]{pennington2014glove}
Jeffrey Pennington, Richard Socher, and Christopher~D. Manning.
\newblock 2014.
\newblock Glove: Global vectors for word representation.
\newblock In {\em Empirical Methods in Natural Language Processing (EMNLP)},
  pages 1532--1543.

\bibitem[\protect\citename{Pustejovsky \bgroup et al.\egroup
  }2003]{pustejovsky2003timeml}
James Pustejovsky, Jos{\'e}~M Castano, Robert Ingria, Roser Sauri, Robert~J
  Gaizauskas, Andrea Setzer, Graham Katz, and Dragomir~R Radev.
\newblock 2003.
\newblock Timeml: Robust specification of event and temporal expressions in
  text.
\newblock {\em New directions in question answering}, 3:28--34.

\bibitem[\protect\citename{Pustejovsky \bgroup et al.\egroup
  }2003a]{pustetal03-IWCS}
James Pustejovsky, Jos\'e Castao, Robert Ingria, Roser Saur\`i, Robert
  Gaizauskas, Andrea Setzer, and Graham Katz.
\newblock 2003a.
\newblock Time{ML}: {R}obust {S}pecification of {E}vent and {T}emporal
  {E}xpressions in {T}ext.
\newblock In {\em Fifth International Workshop on Computational Semantics
  (IWCS-5)}.

\bibitem[\protect\citename{Reimers and
  Gurevych}2017]{reimers-gurevych:2017:EMNLP2017}
Nils Reimers and Iryna Gurevych.
\newblock 2017.
\newblock Reporting score distributions makes a difference: Performance study
  of lstm-networks for sequence tagging.
\newblock In {\em Proceedings of the 2017 Conference on Empirical Methods in
  Natural Language Processing}, pages 338--348, Copenhagen, Denmark, September.
  Association for Computational Linguistics.

\bibitem[\protect\citename{Tonelli \bgroup et al.\egroup }2017]{tonelliimpact}
Sara Tonelli, Alessio~Palmero Aprosio, and Marco Mazzon.
\newblock 2017.
\newblock The impact of phrases on italian lexical simplification.
\newblock In {\em Proceedings of the Fourth Italian Conference on Computational
  Linguistics (CLiC-it 2017)}, Rome, Italy.

\bibitem[\protect\citename{UzZaman \bgroup et al.\egroup
  }2013]{uzzanametal2013}
N.~UzZaman, H.~Llorens, L.~Derczynski, J.~Allen, M.~Verhagen, and
  J.~Pustejovsky.
\newblock 2013.
\newblock {S}em{E}val-2013 task 1: Tempeval-3: Evaluating time expressions,
  events, and temporal relations.
\newblock In {\em Proceedings of {S}em{E}val-2013}, pages 1--9. Association for
  Computational Linguistics, Atlanta, Georgia, USA.

\bibitem[\protect\citename{Verhagen \bgroup et al.\egroup
  }2007]{verhagen-EtAl:2007:SemEval-2007}
M.~Verhagen, R.~Gaizauskas, F.~Schilder, M.~Hepple, G.~Katz, and
  J.~Pustejovsky.
\newblock 2007.
\newblock Sem{E}val-2007 {T}ask 15: {T}emp{E}val {T}emporal {R}elation
  {I}dentification.
\newblock In {\em Proceedings of {S}em{E}val 2007}, pages 75--80, June.

\bibitem[\protect\citename{Verhagen \bgroup et al.\egroup
  }2010]{verhagen2010semeval}
Marc Verhagen, Roser Sauri, Tommaso Caselli, and James Pustejovsky.
\newblock 2010.
\newblock Semeval-2010 task 13: Tempeval-2.
\newblock In {\em Proceedings of the 5th international workshop on semantic
  evaluation}, pages 57--62. Association for Computational Linguistics.

\bibitem[\protect\citename{Vossen \bgroup et al.\egroup
  }2015]{vossen2015storylines}
Piek Vossen, Tommaso Caselli, and Yiota Kontzopoulou.
\newblock 2015.
\newblock Storylines for structuring massive streams of news.
\newblock In {\em Proceedings of the First Workshop on Computing News
  Storylines}, pages 40--49.

\end{thebibliography}

\end{document}